\documentclass[
]{ceurart}

\usepackage{listings}
\newcolumntype{P}[1]{>{\centering\arraybackslash}p{#1}}

\begin{document}

\copyrightyear{2023}
\copyrightclause{Copyright for this paper by its authors.
  Use permitted under Creative Commons License Attribution 4.0
  International (CC BY 4.0).}

\conference{De-Factify 2: 2nd Workshop on Multimodal Fact Checking and Hate Speech Detection, co-located with AAAI 2023}

\title{Overview of Memotion 3: Sentiment and Emotion Analysis of Codemixed Hinglish Memes}

\author[1]{Shreyash Mishra*}[email = shreyash.m19@iiits.in]
 \address[1]{IIIT Sri City, India}

 \author[1]{S Suryavardan*}[email =  suryavardan.s19@iiits.in]

 \author[2]{Megha Chakraborty}
 \address[2]{University of South Carolina, USA}
 
 \author[3]{Parth Patwa}
 \address[3]{UCLA, USA}

 \author[2]{Anku Rani}

\author[4,5]{Aman Chadha\dag}
 \address[4]{Stanford University, USA}
 \address[5]{Amazon AI, USA}
 
 \author[6]{Aishwarya Reganti}
 \address[6]{CMU, USA}

 \author[2]{Amitava Das}[email =  amitava@mailbox.sc.edu]

 \author[2]{Amit Sheth}

 \author[7]{Manoj Chinnakotla}
 \address[7]{Microsoft, USA}

 \author[8]{Asif Ekbal}
 \address[8]{IIT Patna, India}

 \author[9]{Srijan Kumar}
 \address[9]{Georgia Tech, USA}

\renewcommand{\thefootnote}{\fnsymbol{footnote}}
\footnotetext[1]{Equal contribution.}
\footnotetext[2]{Work does not relate to position at Amazon.}
\renewcommand*{\thefootnote}{\arabic{footnote}}
\setcounter{footnote}{0}

\begin{abstract}
Analyzing memes on the internet has emerged as a crucial endeavor due to the impact this multi-modal form of content wields in shaping online discourse. Memes have become a powerful tool for expressing emotions and sentiments, possibly even spreading hate and misinformation, through humor and sarcasm.  In this paper, we present the overview of the Memotion 3 shared task, as part of the DeFactify 2 workshop at AAAI-23. The task released an annotated dataset of Hindi-English code-mixed memes based on their Sentiment (Task A), Emotion (Task B), and Emotion intensity (Task C). Each of these is defined as an individual task and the participants are ranked separately for each task. Over 50 teams registered for the shared task and 5 made final submissions to the test set of the Memotion 3 dataset. CLIP, BERT modifications, ViT etc. were the most popular models among the participants along with approaches such as Student-Teacher model, Fusion, and Ensembling. The best final F1 score for Task A is 34.41, Task B is 79.77 and Task C is 59.82.

\end{abstract}


\maketitle

\section{Introduction} 

    
The term \emph{meme} is derived from the Ancient Greek word \emph{mimema}, meaning \emph{imitated thing}, which comes from the verb \emph{mimeisthai}, meaning \emph{to mimic}. The term was coined by Richard Dawkins, a British evolutionary biologist, in his book \emph{The Selfish Gene} \cite{dawkins_1979}. Dawkins used the concept of a meme to explain the spread of cultural phenomena and ideas, drawing parallels between memes and genes. Dawkins provided examples of memes such as melodies, catchphrases, fashion, and arch-building technology in his book. Interestingly, the word \emph{meme} is a self-describing term, also known as autological, meaning that it is a meme itself.

The reach of online content is vast and has the capacity to reach a massive viewership. Memes propagate through multiple mediums including social media, messaging, apps, and emails. An article on memes quotes "When you plant a fertile meme in my mind, you literally, parasitize my brain" \cite{marwick2013memes}. This quote indicates the impact of memes on a viewer's mind. It spreads to others who find it motivating, offensive, amusing, or relatable in some ways \cite{akhtar}. Memes are used to convey an opinion. When a person comes across a meme that they find relatable, they often share it with others who they think would find relatable.
Different people interpret memes differently which means what can be offensive for one person might not be for other. This difference in interpretation is a result of cultural, gender, and demographic differences.

The memes can also be used to spread offensive and hateful content \cite{multioff}. Identifying and halting the dissemination of hateful memes is an arduous undertaking for both human beings and AI models, as it requires comprehending the intricate subtleties and socio-political contexts that form their interpretation. The deficiencies of current hate speech moderation techniques highlight the pressing need to enhance the effectiveness of automated hate speech detection. Given their subtlety and multi-modal nature, memes pose a more challenging issue to address compared to only text. 

In this paper, we present the findings of the Memotion 3 shared task, where participants were provided with an annotated dataset of 10k memes \cite{mishra2023memotion}, and were tasked with detecting the sentiment, emotion and emotion intensity of the meme. Unlike the previous iteration of memotion \cite{sharma2020semeval,patwa2021findings} which provided English memes, current iteration studies codemixed Hindi-English (Hinglish) memes.

The paper is organized as follows: we describe the related work and the task in section 2 and
3 respectively; Section 4 contains the details of the dataset we collected for Memotion analysis:
Memotion 3.0; followed by a brief description of baseline models and their results in 5. We
conclude and mention some possible future works in section 6.

\section{Related Work}

\textbf{Sentiment and emotion analysis}: 
There has been significant research on sentiment analysis for text over many years \cite{socher-etal-2013-recursive}. Work on sentiment analysis using ML methods like SVM, logistic regression, random forest, XGBoost, k-nearest neighbor has been done in \cite{saad2020opinion, alzyout2021sentiment, prabhakar2019sentiment}. Works which use DL methods include \cite{kokab2022transformer,tesfagergish2022zero,tan2022sentiment}. For detailed surveys on sentiment analysis in social media, please refer to \cite{yue2019survey,chakraborty2020survey}.

The HaHa shared task provides a dataset for humor detection on social media \cite{chiruzzo2019overview}. \citet{ohman2020xed} release an English dataset to detect eight emotions like joy, sadness, disgust etc. A comprehensive survey of textual emotion detection is provided in \cite{acheampong2020text}

\textbf{Hatespeech detection}:
It is important to detect hatespeech to keep social media safe for everyone including minorities. Towards this goal, researchers have curated and released annotated datasets \cite{waseem-hovy-2016-hateful}. The offenseval shared task \cite{zampieri2019semeval} at SemEval 2019 releases an annotated dataset of 14 English tweets to detect the type and target of offensive language. The HatEval shared task releases English and Spanish datasets to detect hate towards women and immigrants. \citet{patwa2021overview} conduct a shared task on Hindi hostile tweets detection. The TRAC workshop series \cite{kumar-etal-2018-benchmarking,trac2-report,kumar-etal-2022-comma} conducts multiple shared tasks to detect aggression and misogyny in English, Hindi, and Bengali datasets. 

Methods to detect hatespeech in text include CNNs and RNNS \cite{gamback2017using,ribeiro2019inf, winter2019know,patwa-etal-2020-hater},   Bert-like models \cite{mazari2023bert,samghabadi2020aggression,risch-krestel-2020-bagging}, incorporating linguistic characteristics \cite{nagar2023towards} etc.

\textbf{Codemixed Language Processing} : Codemixed language processing is a challenging task because the informal mixing of 2 or more languages and the proliferation of unique number of ways to write the same word \cite{Laddha_Hanoosh_Mukherjee_Patwa_Narang_2020,Laddha_Hanoosh_Mukherjee_Patwa_Narang_2022}. The Sentimix task \cite{patwa-etal-2020-semeval} at semeval 2020 focused on sentiment analysis of Hinglish and Spanish-English tweets. \cite{chakravarthi-etal-2021-findings-shared} organized a shared task on detecting offense in 3 codemixed dravidian languages - Tamil \cite{chakravarthi-etal-2020-corpus}, Malayalam\cite{chakravarthi-etal-2020-sentiment} and Kannada\cite{hande-etal-2020-kancmd}. Methods explored to tackle codemixing include graph convolutional netowrks \cite{dowlagar2021graph}, BERT based models \cite{risch2019hpidedis,tula-etal-2021-bitions}, modifying loss function \cite{Tula2022cmi,ma-etal-2020-xlp} or positional embeddings \cite{Ali-Kandukuri-2022} to incorporate codemixing etc.

\textbf{Multimodal analysis }: Although most of the existing research focuses on unimodal (text) analysis, the use of multi-modal content likes memes and videos is fast increasing. Multimodal datatsets having text and image, or videos are useful for tasks like image captioning, hatespeech detection, emotion analysis in videos, sentiment analysis \cite{Hu-2018,jha-etal-2021-image2tweet,gomez2019exploring,bagher-zadeh-etal-2018-multimodal} etc among other tasks. 
The Hateful Memes Challenge dataset \cite{kiela2020hateful} and the multioff dataset \cite{multioff} address hatespeech and offense detection in memes. However, they are binary classification task and the memes are in English whereas memotion 3 has multi-class and multi-label tasks on code-mixed data. 
There have been very few works on code-mixed meme analysis. \cite{suryawanshi-chakravarthi-2021-tamil} conduct a shared task to detect trolling in Tamil codemixed memes whereas \cite{hossain-etal-2022-mute} release a dataset to detect hate in codemixed Bengali memes.
Popular methods for multimodal learning include image-text joint embeddings \cite{Xie-2021-JE,krishna2023imaginator,Gunti-JE-2022} and transformer based models \cite{lu2019vilbert,li2019visualbert}.

\textbf{Previous iteration of Memotion} : Memotion 1 \cite{sharma2020semeval} and Memotion 2\cite{patwa2021findings} shared task released datasets of 10k memes each \cite{sharma2020semeval,ramamoorthy2022memotion}. These datasets were annotated on the same tasks as Memotion 3. However, both these datasets only focus on English memes, where as in memotion 3, we focus on Hinglish codemixed memes. Methods like ensembling \cite{phan2022little,morishita-etal-2020-hitachi-semeval-2020,guo-etal-2020-guoym} and bert-like models \cite{vlad2020upb,nguyen2022hcilab,lee2020amazon} were common across memotion 1 and Memotion 2.

\section{Task Details}

\subsection{Tasks}

Memotion 3 is the 3rd iteration of the Memotion task. The challenge consists of three sub-tasks: 
\begin{enumerate}
    \item {\textbf{Task A - Sentiment analysis of memes}: Given a meme, the system is supposed to classify the meme's sentiment as positive, negative or neutral.}
    \item {\textbf{Task B - Overall emotion analysis of memes}: This task's goal is to pinpoint certain emotions connected to a given meme. Whether a meme is humorous, sarcastic, offensive, or motivating should be indicated by the system. There are multiple categories in which a meme can fit.}
    \item {\textbf{Task C - Classifying the intensity of meme emotions}: The task is to determine the degree to which a particular emotion is being expressed. The ranking of these emotions is as follows:

    \begin{enumerate}
        \item Humour: Not funny, funny, very funny and hilarious
        \item Sarcasm: Not Sarcastic, little sarcastic, very sarcastic and extremely sarcastic
        \item Offensive: Not offensive, slightly offensive, very offensive and hateful offensive
        \item Motivation: Not motivational, motivational
    \end{enumerate}
    }
    
\end{enumerate}

\subsection{Dataset}
The tasks were conducted on the Memotion 3 dataset \cite{mishra2023memotion}. It consists of Hindi-English codemixed memes, which were collected from selenium based web crawler and they were gathered from various public platforms like Reddit, Google Images, etc and annotated  manually. The dataset consists of 10,000 meme images divided into a train-val-test split of 7000-1500-1500. Each meme is annotated for its sentiment, emotion and intensity of emotion. Images also have their corresponding OCR text extracted with the help of Google
Vision APIs and their respective URLs. For more details of the  dataset, please refer to \cite{mishra2023memotion}.

\subsection{Evaluation}

As mentioned previously, there are three tasks. Scoring is done for each task separately, and separate leaderboards are generated. For each task, we use weighted average F1 score to measure the performance of a model. The participants had access to only train and validation set. They were asked to submit a maximum of 3 submissions on the test set for each task, the best of which was selected as part of the leaderboard.

\subsection{Baselines}

For multi-modal data, it is crucial to take into account both the visual and textual properties, particularly in the case of memes where the context can only be recorded using a mix of both elements. For textual features, we employ a multilingual form of BERT, namely Hinglish-BERT (\texttt{BERT-base-multilingual-cased}) \cite{bhange2020hinglishnlp}, which is tuned on Hinglish data. The trained Vision Transformer (ViT) model provides the visual properties \cite{dosovitskiy2020imagevit}. The Hinglish-BERT embedding is concatenated with the pooled output from the ViT model. After passing through an MLP, the combined features are then categorised in a final classification layer. For more details about the baseline, please refer \cite{mishra2023memotion}.
\section{Participating systems}
There were 47 team registrations for the task in the Memotion 3.0 task page, of which 5 teams made submissions for the final test set of the dataset. The results for all three tasks are given in the following section and an overview of the 4 teams that presented their description papers are provided below.

\textbf{wentaorub} \cite{wentaorub} use CLIP \cite{radford2021learning} for individual text and image embeddings, before concatenating them and passing them through multi-headed attention layers for classification. They also use the OSCAR \cite{li2020oscar} model in this approach and an ensemble of their models are used for the final submission. Datasets such as Facebook Hateful memes \cite{kiela2020hateful}, MMHS150k \cite{gomez2019exploring} etc. are used for pre-training. This architecture helped them achieve the best performance in Task B and C.

\textbf{NYCU\_TWO} \cite{NYCU_TWO} propose a two model pipelines, namely Coopoerative Teaching Model (CTM) for task A and Cascaded Emotion Classifier (CEC) for task B and C. A fusion of multi-modal embeddings from pre-trained Swin-Transformer \cite{liu2021swin} and CLIP are passed to the CTM and CEC pipelines. CEC helps leverage task C predictions for task B by jointly training the model. This team attained the best results in 3 out of the 4 labels in Task B.

\textbf{NUAA-QMUL-AIIT} \cite{NUAA} refer to their approach as Squeeze-and-Excitation Fusion or SEFusion. The textual features from pre-trained RoBERTa \cite{liu2019roberta} and visual features from CLIP-ViT \cite{zhai2020largescale} are  fused to obtain multi-modal embeddings. The fusion is the SEFusion module, which uses a learned activation of the squeezed features, allowing for weighted fusion of multi-modal embeddings. This approach led to NUAA-QMUL-AIIT being the 1st ranked team in Task A.

\textbf{CUFE}  used a LightGBM \cite{lightgbm} classifier for classification on every individual emotion or label in all Tasks. The inputs to the classifier were pre-trained Hinglish-DistilBERT for text embeddings and ResNet18 \cite{resnet} for image embeddings. Other features, such as occurrence of characters, word count etc. from text and number of faces in the memes (extracted using Facenet’s \cite{Schroff2015} multitask cascaded CNN), were also used. CUFE obtained the highest score in 2 labels in Task B and 1 label in Task C.
\section{ Results}

\begin{table}[t!]
    \centering
    \begin{tabular}{P{1cm}p{4cm}P{3cm}}
    \toprule
    Rank & Team & F1 - Scores \\
    \toprule
    1 & \textbf{NUAA-QMUL-AIIT} & \textbf{34.41\%} \\
    2 & NYCU\_TWO & 34.20\% \\
    3 & CUFE & 33.77\% \\
    4 & CSECU-DSG & 33.34\% \\
    5 & Baseline & 33.28\% \\
    6 & wentaorub & 32.88\% \\
    \bottomrule
    \end{tabular}
    \caption{Leaderboard of teams on Task A: Sentiment Analysis.}
    \label{tab:taska_results}
\end{table}

The performance in Task A i.e. sentiment classification is presented in Table \ref{tab:taska_results}. Our proposed baseline achieved a score of 33.28\%. Out of the five final submissions, four teams managed to surpass the baseline. The top two teams, namely, NUAA-QMUL-AIIT \cite{NUAA} and NYCU\_TWO \cite{NYCU_TWO} improved on the baseline by 1.1\% and 0.9\% respectively. The difference in F1 scores for this task is minimal across all participants.

\begin{table}[h]
    \centering
    \begin{tabular}{P{0.75cm}p{2.8cm}P{1.5cm}P{1.5cm}P{1.5cm}P{1.5cm}P{1.5cm}}
    \toprule
    Rank & Team & \multicolumn{5}{c}{F1 - Scores}\\
    & & H & S & O & M & Overall\\
    \toprule
    1 & wentorub & 88.91 & 86.74 & 48.99 & \textbf{94.44} & \textbf{79.77}\\
    2 & NYCU\_TWO & \textbf{89.04} & \textbf{86.91} & 43.17 & \textbf{94.44} & 78.39\\
    3 & NUAA-QMUL-AIIT & 87.06 & 77.97 & 50.72 & \textbf{94.44} & 77.55\\
    4 & BASELINE & 84.55 & 74.82 & 48.84 & 90.78 & 74.74\\
    5 & CUFE & 82.26 & \textbf{86.91} & \textbf{50.78} & 77.60 & 74.39\\
    6 & CSECU-DSG & 88.64 & 86.30 & 49.32 & 64.79 & 72.26\\
    \bottomrule
    \end{tabular}
    \caption{Leaderboard of teams on Task B: Emotion analysis \{H:Humor, S:Sarcasm, O:Offense, M:Motivation\}. All the scores are in percentage. The teams are ranked by their average F1 scores (overall) across all the four emotions. Motivation emotion is the easiest to detect while offense is the most difficult to detect.  }
    \label{tab:taskb_results}
\end{table}

Table \ref{tab:taskb_results} shows the leaderboard for Task B. Three teams out-performed the baseline for this task and top performing team \texttt{wentaorub} \cite{wentaorub} improves on the final score by 5.4\%. Team CUFE  performs lower than the baseline score by 3.4\%, however, they present the highest score on the Offensive sub-task and on the Sarcasm sub-task, jointly with NYCU\_TWO \cite{NYCU_TWO}. \texttt{wentaorub} \cite{wentaorub}, \texttt{NYCU\_TWO}, and \texttt{NUAA-QMUL-AIIT} achieved the same score on Motivation, which is 4\% higher than the baseline. \texttt{NYCU\_TWO} also scored the highest in the Humour sub-task, improving on the baseline by 4.5\%. Based on these F1 scores, we can deduce that the Motivation class is the easiest to detect, similar to Memotion 2.0 \cite{ramamoorthy2022memotion}, despite some teams performing poorly on this class. However, in this iteration, the Offensive class is the hardest to detect, instead of the Sarcasm class. No single team performs the best on all the classes. 

\begin{table}[h]
    \centering
    \begin{tabular}{P{0.75cm}p{2.8cm}P{1.5cm}P{1.5cm}P{1.5cm}P{1.5cm}P{1.5cm}}
    \toprule
    Rank & Team & \multicolumn{5}{c}{F1 - Scores}\\
    & & H & S & O & M & Overall\\
    \toprule
    1 & wentaorub & \textbf{48.37} & 50.65 & \textbf{45.82} & \textbf{94.43} & \textbf{59.82} \\
    \hline
    2 & NUAA-QMUL-AIIT & 46.34 & 44.29 & 43.17 & \textbf{94.43} & 57.06 \\
    \hline
    3 & CSECU-DSG & 42.54 & 34.41 & 45.68 & 93.26 & 53.97 \\
    \hline
    4 & NYCU\_TWO & 43.28 & 45.80 & 44.67 & 80.57 & 53.58 \\
    \hline
    5 & CUFE & 41.66 & \textbf{51.07} & 42.02 & 77.60 & 53.09 \\
    \hline
    6 & BASELINE & 43.03 & 32.89 & 42.40 & 90.78 & 52.27 \\
    \bottomrule
    \end{tabular}
    \caption{Leaderboard of teams on Task C: Emotion intensity detection \{H:Humor, S:Sarcasm, O:Offense, M:Motivation\}. All scores are in percentage. The teams are ranked by average F1 scores (overall) across all the four emotions. Intensity of sarcasm is by far the most difficult to detect. Motivation has only two intensities as opposed to four intensity levels for other emotions.}
    \label{tab:taskc_results}
\end{table}

As shown in the leaderboard for Task C in Table \ref{tab:taskc_results}, all the participating teams outperform the baseline in this task. The minimum improvement in F1 score is 0.8\% by team CUFE  and the maximum improvement in the final score is 7.5\% by \texttt{wentaorub} \cite{wentaorub}. The submission by \texttt{wentaorub} exceeds all other teams in the sub-tasks Humour, Offensive and Motivation. \texttt{NUAA-QMUL-AIIT} \cite{NUAA} matched the highest score of \texttt{wentaorub} in the Motivation category, while CUFE  achieved the highest score in the Sarcasm sub-task. The Motivation class has significantly higher F1 scores than other classes, due to it having only 2 intensities whereas other classes have 4 intensities each. The best overall score is only 59.82\%, which shows the difficulty of the task. 

\begin{figure}[h]
    \centering
    \begin{tabular}{ccc}
    \includegraphics[width = 1.6in, height = 1.25in]{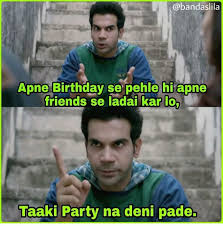} &
    \includegraphics[width = 1.6in, height = 1.25in]{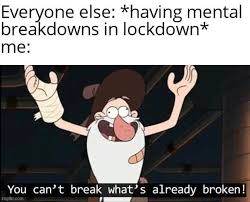} &
    \includegraphics[width = 1.6in, height = 1.25in]{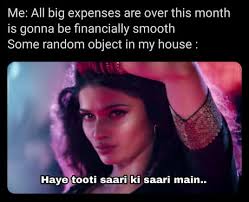} \\
    \includegraphics[width = 1.6in, height = 1.25in]{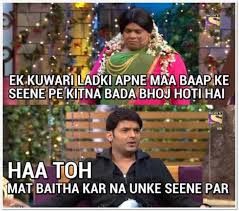} &
    \includegraphics[width = 1.6in, height = 1.25in]{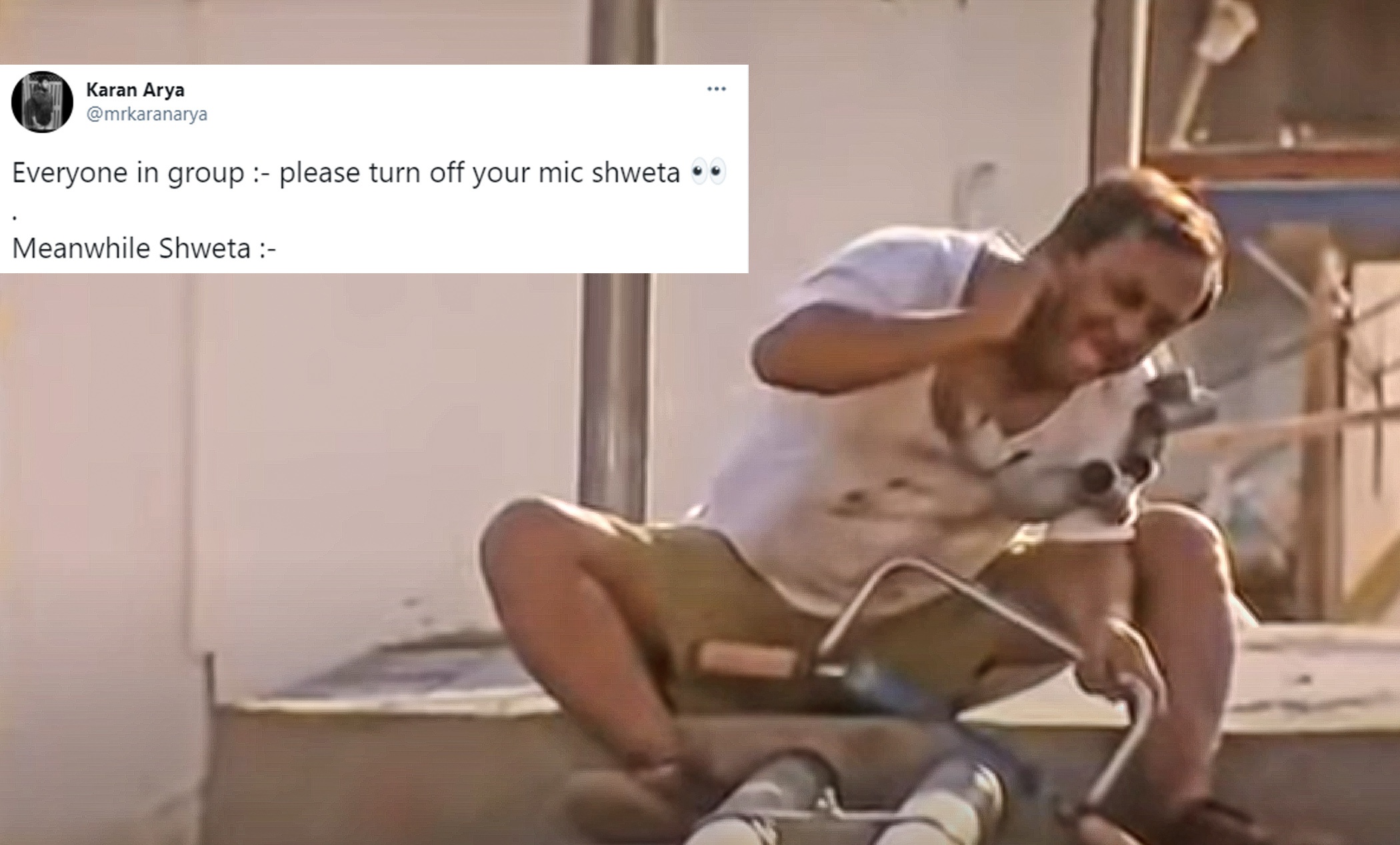} &
    \includegraphics[width = 1.6in, height = 1.25in]
    {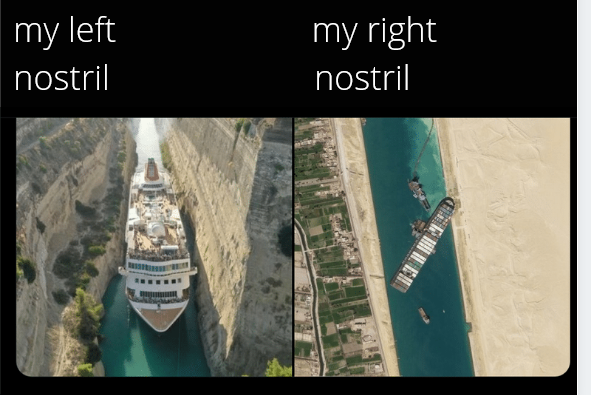} \\
    \end{tabular}
    \caption{Some samples for Task A from the 121 memes that were predicted incorrectly by all participating teams. Over half of the mis-classified memes are true negatives. }
        \label{fig:wrongsA}
\end{figure}

\begin{figure}[h]
    \centering
    \begin{tabular}{ccc}
    \includegraphics[width = 1.6in, height = 1.25in]{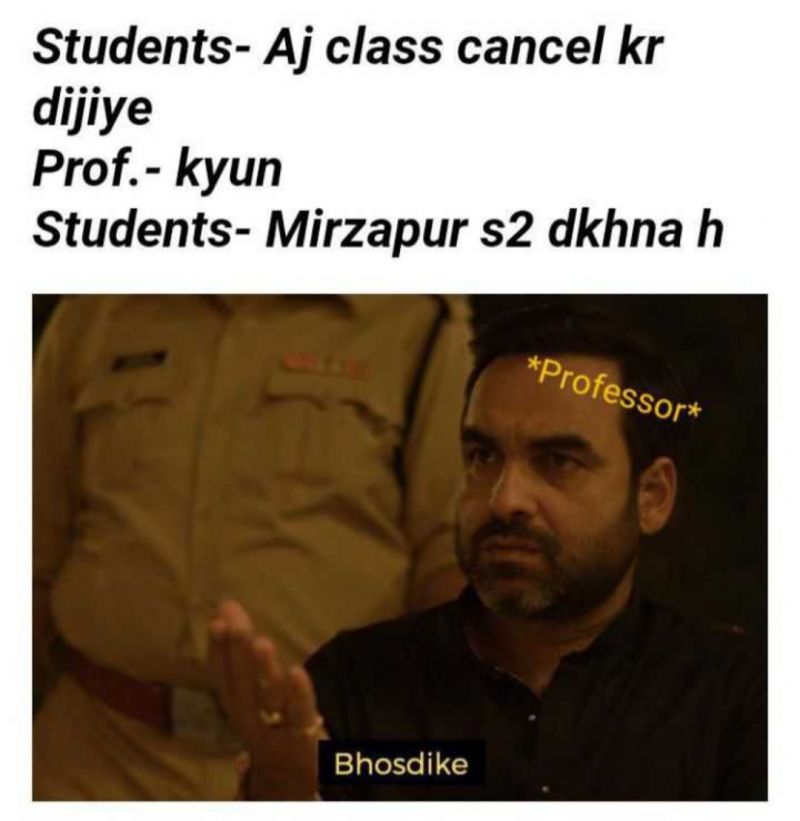} &
    \includegraphics[width = 1.6in, height = 1.25in]{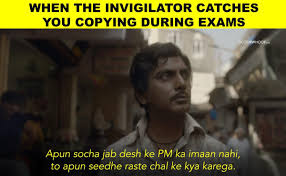} &
    \includegraphics[width = 1.6in, height = 1.25in]{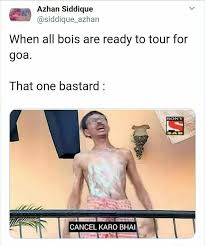} \\
    \includegraphics[width = 1.6in, height = 1.25in]{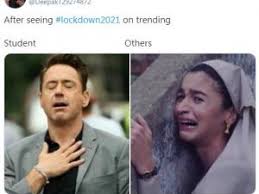} &
    \includegraphics[width = 1.6in, height = 1.25in]{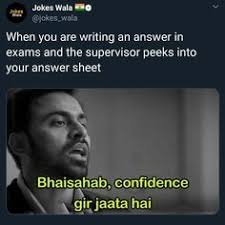} &
    \includegraphics[width = 1.6in, height = 1.25in]
    {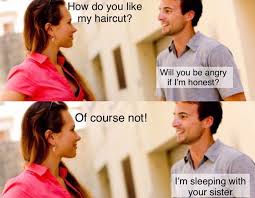} \\
    \end{tabular}
    \caption{242 memes from the test set were predicted incorrectly by all participants. Most of such memes are code-mixed and related to humor or sarcasm. }
        \label{fig:wrongsB}
\end{figure}

\begin{figure}[h]
    \centering
    \begin{tabular}{ccc}
    \includegraphics[width = 1.6in, height = 1.25in]{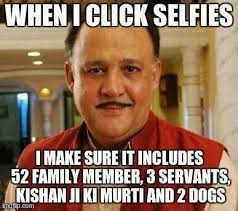} &
    \includegraphics[width = 1.6in, height = 1.25in]{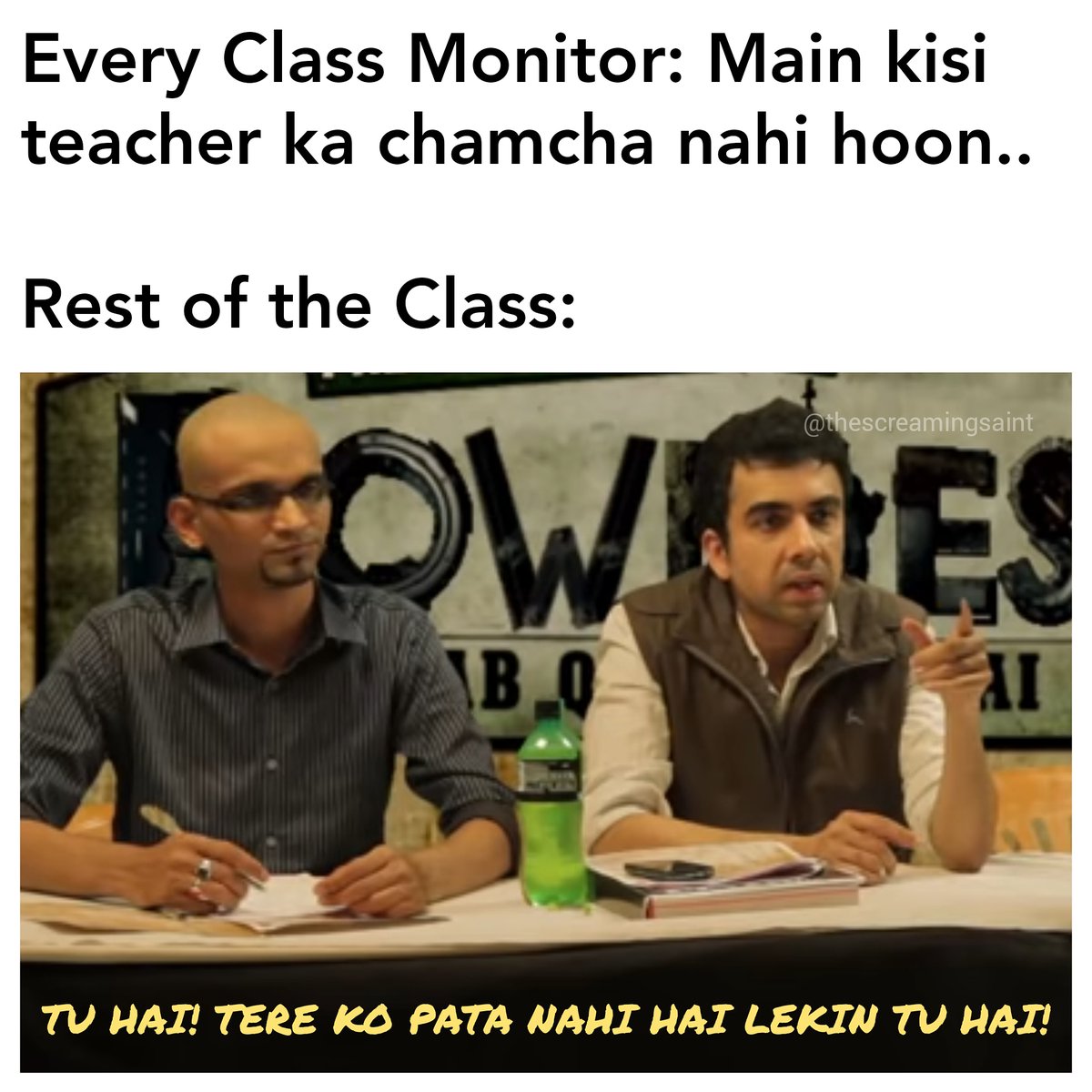} &
    \includegraphics[width = 1.6in, height = 1.25in]{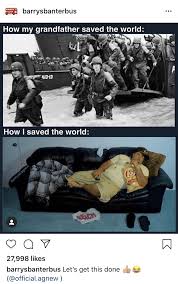} \\
    \includegraphics[width = 1.6in, height = 1.25in]{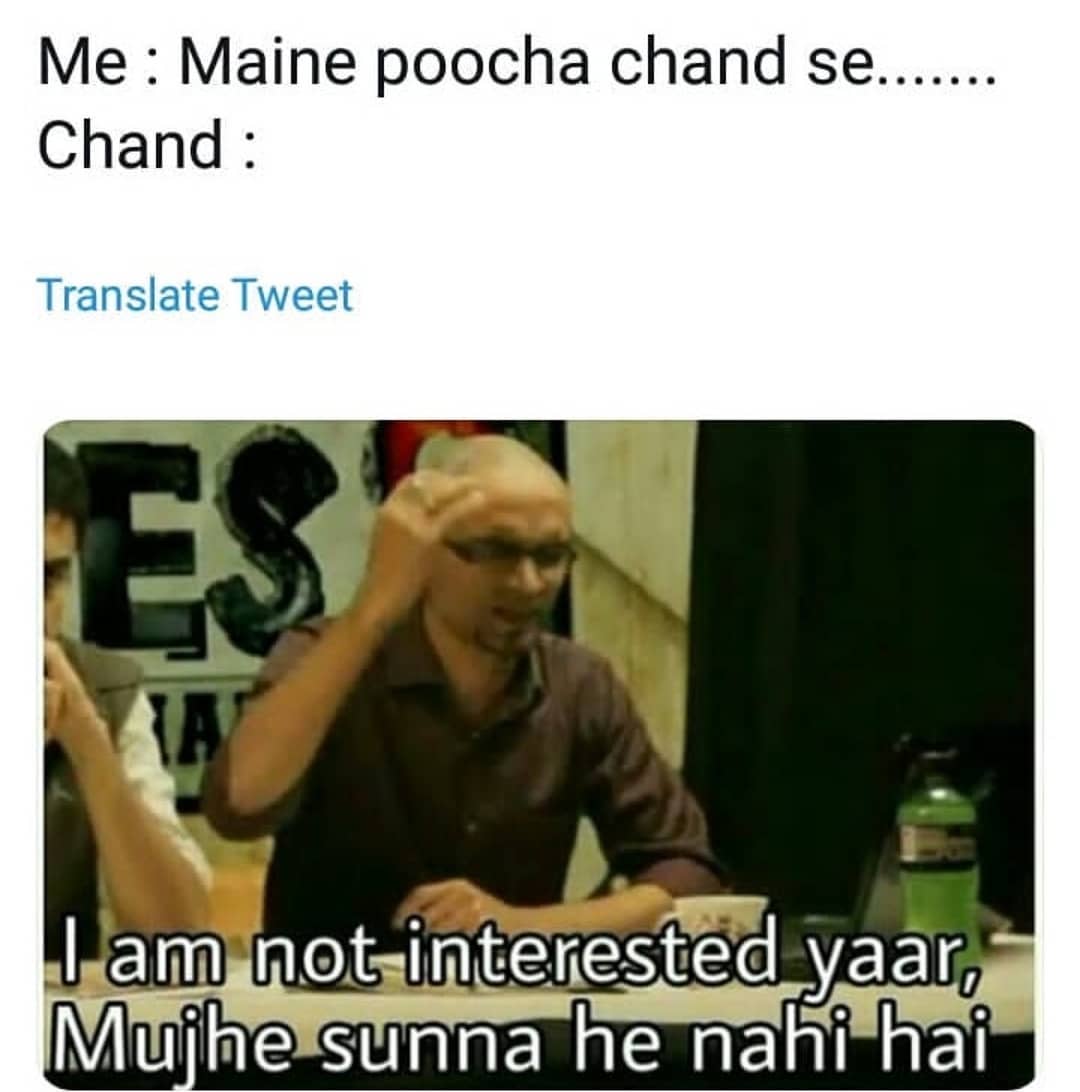} &
    \includegraphics[width = 1.6in, height = 1.25in]{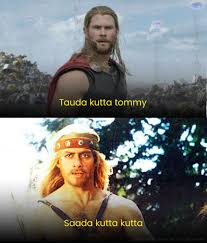} &
    \includegraphics[width = 1.6in, height = 1.25in]
    {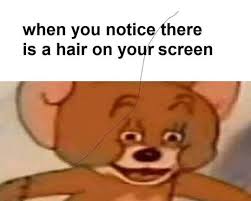} \\
    \end{tabular}
    \caption{For Task C, none of the teams made correct predictions on over 400 memes from the test set. A large share of these memes is code-mixed, thus indicating the added challenge to the task.}
        \label{fig:wrongsC}
\end{figure}

A major observation when comparing performances with the previous iteration of the task \cite{patwa2021findings} is that the performance on Task A is much worse in Memotion 3. Overall Scores in Tasks B and C have remained about the same compared to Memotion 2.

For each individual task, we analyze the memes from the test set that all participants made wrong predictions on. For Task A, there are 121 memes where all participants mis-classified the label, out of which 66 memes were true negative sentiment followed closely by true positive memes. For task B, there are 231 such memes, majority of which belong to "humor" and "sarcasm" class. Finally, for Task C, 421 memes are mis-classified by all the systems - most memes mis-classified by all teams are "Very Funny", "Very Sarcastic", "Slightly Offensive". Some such examples for Task A, B and C are shown in  figures \ref{fig:wrongsA}, \ref{fig:wrongsB}, and \ref{fig:wrongsC} respectively. Further, we note that most of such difficult examples have code-mixed text. 

As for the overall performances, only two teams - \texttt{NUAA-QMUL-AIIT} and \texttt{NYCU\_TWO} \cite{NUAA, NYCU_TWO} - perform better than the baseline in all tasks.

\section{Conclusion and Future Work}
In this paper, we summarize the approaches used by the participants for the Memotion 3 task and analyze the results.  Due to the multi-modal nature of the dataset, all teams use a pre-trained image and text embedding models. However, each team presents a novel model pipeline. The highest scores achieved in Task A, Task B and Task C of Memotion 3.0 are 34.41\%, 79.77\% and 59.82\% respectively, which shows there is significant room for improvement. On analysis of the results and the mis-classified examples on the test set, we find that "Sarcasm" and "Humour" are difficult to identify, especially in code-mixed memes.

While we address Hind-English code-mixed memes in this paper, future work could include exploring other languages/language pairs. A unified baseline model to analyze memes in multiple languages could also be an interesting possibility.

\bibliography{references}

\end{document}